# How Emotional Mechanism Helps Episodic Learning in a Cognitive Agent

Usef Faghihi, Philippe Fournier-Viger
Student
Université du Québec à Montréal
201, avenue du Président-Kennedy,
Montréal, Canada, H2X 3Y7
1-(514)-987-3000 #8922
{faghihi.usef, fournier_viger.philippe,
}@courrier.uqam.ca

Roger Nkambou, Pierre Poirier
Professor
Université du Québec à Montréal
201, avenue du Président-Kennedy,
Montréal, Canada, H2X 3Y7
1-(514)-987-3000 #8922
{nkambou.roger, poirier.pierre}@uqam.ca

André Mayers
Professor
Université de Sherbrooke
2500, boul. Université
Sherbrooke, Canada, J1K 2R1
1-(819)-821-8000 #2041
andre.mayers@usherbrooke.ca

## ABSTRACT
In this paper we propose the CTS (Concious Tutoring System) technology, a biologically plausible cognitive agent based on human brain functions.This agent is capable of learning and remembering events and any related information such as corresponding procedures, stimuli and their emotional valences. Our proposed episodic memory and episodic learning mechanism are closer to the current multiple-trace theory in neuroscience, because they are inspired by it [5] contrary to other mechanisms that are incorporated in cognitive agents. This is because in our model emotions play a role in the encoding and remembering of events. This allows the agent to improve its behavior by remembering previously selected behaviors which are influenced by its emotional mechanism. Moreover, the architecture incorporates a realistic memory consolidation process based on a data mining algorithm.

## Categories and Subject Descriptors
I.2.11 [**Distributed Artificial Intelligence**]: Intelligent agents; I.2.0 [**General**]: Cognitive simulation

## 1. INTRODUCTION
As is the case in humans, memory is very important for cognitive agents. Episodic memory is one particularly important form of memory [1]. It allows an agent to adapt its behavior to the environment according to previous experiences. In humans, the memory of *what*, *where* and *when*, known as episodic memory, is influenced directly or indirectly by the amygdala, which play a major role in emotional processes [2]. Many have attempted to incorporate episodic memory and learning mechanisms in cognitive agents and cognitive architectures, yet none have included a role for emotions in the episodic learning and retrieval processes.

Recently, studies have demonstrated the role of the hippocampus and its influences on episodic memory consolidation in the human brain. These suggest that a fast form of learning occurs first in the hippocampus. The information is then transferred, via a slower process, to various cortical areas.

Two models are suggested in neuroscience for the human consolidation phase [5], (1) the standard consolidation theory and (2) the multiple-trace theory. The standard consolidation theory holds a hippocampus-independent view of event encoding. It posits that the hippocampus performs a fast interpretation and learning of a given concept or event. In the transfer phase, indirect connections are thought to be created between the hippocampus and various neurons in the cortex. The hippocampus then distributes these memory traces to the cortex. Importantly, in this model, the cortical neurons representing events create direct connections between themselves and gradually become independent of the hippocampus.

The multiple-trace theory, on the other hand, holds a hippocampus-dependent view of event encoding. According to this theory, every time an event causes memory reactivation, a new trace for the activated memory is created in the hippocampus. Memory consolidation occurs through the reoccurring loops of episodic memory traces in the hippocampus and the construction of semantic memory traces in the cortex. Thus, the cortical neurons continue to rely on the hippocampus throughout encoding.

Emotions affect different types of memory and enhance learning in humans [3]. Scientists now know that our brains automatically create emotional valences of events in episodic memory. In fact, recent neuroimaging studies [2] showed a relation between the activation of the anterior temporal lobe and emotional memory retrieval. They demonstrated that the brain regions known to be involved in episodic memory retrieval are also useful in the retrieval of emotional episodic memories. Because emotions and episodic memory play two complementary roles in learning and the retrieval phase in the human brain, we believe that both must be included in cognitive architectures.

CTS is inspired by the latest neurobiology and neuropsychology theories of human brain function (see figure 1). It is an agent designed to provide assistance during training in virtual learning environments. In this work, it is applied to a tutoring system in order to provide assistance to astronauts learning how to manipulate Canadarm2, the robotic telemanipulator attached to the International Space Station (ISS). Our team has added an emotional mechanism into the CTS (Conscious Tutoring System) [4] architecture.

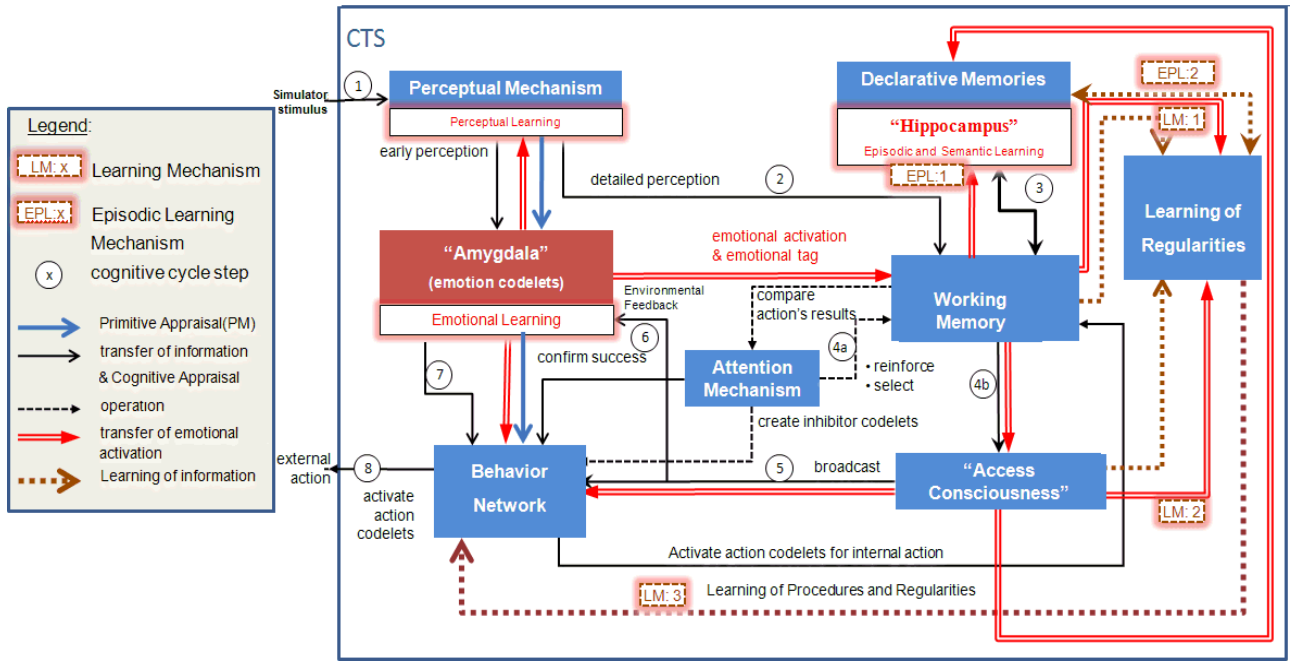

**Figure 1 CTS Architecture**

We here explain the crucial role episodic memory (or pseudo-hippocampus) plays in our model's episodic learning mechanisms associated with emotion. These mechanisms consist of the encoding of any given information coupled with its assigned emotional valence, and the encoding of the agent's actions.

The episodic memory or "pseudo-hippocampus" is composed of two main mechanisms called "memory consolidation" and "episodic learning" [5]. These intervene in the memorization and the retrieval phases of the events in CTS's memory architecture respectively. The memorization phase of the CTS architecture includes emotional valences ascribed to ongoing events by the Emotional Mechanism [6]. Emotional valences are here taken to be CTS's memorization of valenced reactions to given emotional situations (stimuli) as described in the OCC model [7]. All events sequences are distributed and then stored in different memories in CTS. The memorization phase also includes the process of memory consolidation [5]. This process constantly extracts temporal regularities from all past episodes to form a procedural memory. This process is very important because as a cognitive agent, CTS receives a huge amount of data which is temporally related to its environment. Moreover, much communication takes place between the different parts of the system. This produces a large amount of data during each cognitive cycle which accumulates. In order to be used in decision making, it needs to be consolidated into a smaller form. In our architecture, this is achieved through sequential pattern mining [8], for it is an efficient knowledge discovery technique that is widely used in computer science to find frequent temporal patterns among sequences of symbols to face huge amount of data, a common situation with CTS. This provides a functionally plausible memory consolidation model. These sequential patterns are useful in the retrieval (remembering) phase to adapt CTS's behavior to past experiences. In the retrieval phase, a retrieval cue is introduced to all sequential patterns previously created by the system, making them active, each according to its likeness to the cue. The information sequence activated in parallel thus reinforces the cue's content.

Of course, this generic architecture of emotion and learning could be implemented in any cognitive or real-time agent. In our specific case it helps to sustain better tutorial interventions. In the present paper, we will begin by a brief review of the existing work concerning episodic learning in cognitive agents. We will then propose our new architecture combining elements of the Emotional Mechanism (EM) and episodic memory. Finally, we will present results from our experiments with this cognitive agent.

## 2. A COMPARISON BETWEEN EPISODIC MEMORY IMPLEMENTED IN DIFFERENT COGNITIVE AGENTS

Various models of episodic memory have been proposed and implemented, for example LIDA [9]. They have attempted to implement episodic learning [9] but none have succeeded. To begin, it must be noted that Sidney et al. have not performed any empirical experiments about episodic learning on their proposed agent LIDA. Moreover, in their agent, the Episodic and Semantic Memories are separated and have totally different structures - this impedes a joint memory consolidation. In humans however, episodic and semantic memories activate overlapping brain regions [5]. In this way, episodic and semantic memories consolidate each other during the learning of a new event and during the remembering phase. These are ignored in LIDA's architecture. It is also unclear how LIDA encodes time for a given episode in its system. Lastly, an event may be associated to a procedure. In their paper they do not address how this is possible.

In the well-known ACT-R model, there is no explicit episodic memory. Instead, episodic events are encoded in the declarative memory as chunks, just like declarative information. During the recall, beside the activation provided by the context, a base level activation function is used for each chunk to calculate the probability of being retrieved and the speed of its retrieval. Basically, the activation is calculated based on the time elapsed since the last occurrence of the chunk in Working Memory (WM) and the number of times that the chunk was recalled. Because chunk activation decreases rapidly over time, after a short while, the frequency of chunk use becomes the most decisive feature for determining recall. Thus, ACT-R cannot recall information in a temporal context, and this induces abnormal behavior [19]. In addition, since ACT-R has no emotions, these cannot be taken into account during episodic memorization and retrieval.

## 3. EPISODIC MEMORY IN CTS ARCHITECTURE

Our generic agent is an extension of the generic cognitive architecture underlying it. It relies on the functional "*consciousness*" [11] mechanism for much of its operations. Its functional architecture is inspired from LIDA's [11], with some differences and some extensions.

In order to assist the astronaut in learning how to manipulate the Canadarm2, CTS considers a multitude of elements: the current simulator state (position and configuration of Canadarm2 as depicts Figure 2) [17], the potential dangers such as an imminent collision, the quality of the learner's actions, the learner's knowledge state and preferences, his learning objectives, and so on. Like LIDA, CTS bears some functional similarities with the physiology of the nervous system, and its cognitive cycle incorporates a more detailed version of the traditional Perception-Reasoning-Action cycle. Its modules communicate with one another by contributing information to its Working Memory through information codelets[1]. These travel back to modules and other unconscious resources through loops of "conscious publications" that broadcast only the most important, urgent, or relevant information (as suggested by Baars [12]).

In CTS, a "*pseudo-amygdala*" is responsible for emotional reactions [6]. In this work, we have added also a pseudo-hippocampus (PH) which mediates all declarative memory in CTS [5]. A very good interaction between emotional and episodic mechanism is crucial for the memorization and retrieval phase.

In CTS, the cognitive cycle (see figure 1) begins with perception codelets[13] performing collective interpretations. These result in a percept, which is temporarily stored in the Perception Network. The percept, which is constituted of the active nodes of the Perceptual Network (PN), enters Working Memory as a single network of codelets. These codelets create or reinforce associations with other already present codelets and create a coalition of information codelets (these may include emotional content and receive additional positive/negative energies from the CTS pseudo-amygdala). The emotional codelets situated in the CTS pseudo-amygdala inspect each coalition's informational content, and infuse it with a level of activation proportional to its emotional valuation. This increases their likeliness to be selected for various memories' (episodic and semantic) retrieval phase and after retrieval phase. This information is considered as an elementary indication for the system.

This information is submitted to the CTS *pseudo-hippocampus* and the stored traces (indexed information) from the pseudo-hippocampus [5] are activated to find local relationships. Corresponding information from different memories are automatically retrieved. The indexed information considered as events for pseudo-hippocampus system includes: the emotional valences of each codelet, the current and previous activation of each codelet, the last cognitive cycle in which codelets called to WM, and the current cognitive cycle.

The *pseudo-hippocampus* plays two important roles in CTS's cognitive cycles, (1) during memory consolidation and (2) during information retrieval. These are discussed below.

The *pseudo-hippocampus* is essential during (1) memory consolidation, for it retrieves information from the different memories (see multiple-trace theory [5]) to create a centralized sequence of events. An event is defined as the information codelets contained in the WM with their activation value and the emotional valences at the current cognitive cycle. The sequence of events is a list of events that occurred including the current event and residual information from previous cognitive cycles in WM. The retrieved indexed information by pseudo-hippocampus contains codelet links with other codelets. Each time they enter WM, this information is updated according to the links made between these codelets and others in the coalition present in WM. At this point, attention codelets observe WM to select urgent, decisive or specific information and try to bring it into consciousness (Franklin and Patterson, 2006) broadcast mechanism, competing with other coalitions in WM. The Attention mechanism spots the most energetic coalition in WM and submits it to the CTS broadcast mechanism. With this broadcast, any subsystem that recognizes the information (appropriate module or team of codelets) in any part of the system may react to it.

The pseudo-hippocampus is also necessary during (2) memory retrieval. After information is broadcasted, in the reasoning phase, it must retrieve the past frequently reappearing information best matching the current information resident in WM. The pseudo-hippocampus extracts frequent partial or complete event sequences (episodic patterns) from the list of special events previously consolidated. This may invoke a stream of behaviors related to the current event, with activation passing through the links between them. These could be considered partial or complete (procedures) of actions. It is important to note that the LIDA architecture ignored this association between an event and its related procedures [9]. In our model, this process can be considered as a recalculation and may strengthen or weaken the base-level energy and links made between different codelets in WM, according to previous similar event sequences. This can be considered as a learning process in this cognitive cycle, which we detail in the next section. The entire processes explained above may create reflection loops that further pursue a line of reasoning through consecutive participation of various sub-systems in cycles of selection-broadcast-reaction. Up to now in the CTS architecture, the Behavior Network (BN), reacting to broadcast, planned actions and monitored frequent partial or complete events sequences [5](page 370). With our revised CTS model, the pseudo-hippocampus now also proposes an action through an emergent selection process, and decides upon the most appropriate to adopt.

---

[1] Based on Hofstadter *et al.'s* idea, a *codelet* is a very simple agent, "a small piece of code that is specialized for some comparatively simple task". Implementing Baars theory's *simple processors*, codelets do much of the processing in the architecture.

## 4. CTS's EPISODIC LEARNING

In this section we explain how Episodic Learning is implemented and how different learning collaborates in the CTS's architecture. We then explain how our Episodic Memory consolidation model and Episodic learning process occur in CTS's architecture.

### 4.1 Episodic and Procedural Learning in CTS

We would like to explain very briefly CTS's Behavior Network [14] functionality before explaining Episodic Learning (EPL).

Aside from the learning of environmental regularities, procedural learning is also implemented in the CTS architecture. CTS's procedural memory mechanism is influenced by episodic learning mechanism interventions. Procedural memory is a CTS behavioral network of partial plans that analyses the context to decide what to do, and which behavior to set off. This structure is linked to the latent knowledge of how to do things in the form of inactive codelets. Each behavior node, just as a codelet, has a base-level activation, which can increase or decrease. Until it is selected for execution, a behavior node goes back-and-forth gaining and losing energy from various sources in the BN. It must be noted that the links between the nodes are concerned with energy as well. Procedural learning starts when external stimuli are interpreted by CTS's perceptual mechanism and written into WM, where they may then be chosen by the Attention mechanism to be presented to consciousness. The broadcasted information may either assert preconditions for the initiation of a behavior in BN, or cause reactions in another part of the system, which then creates the necessary preconditions to fire a behavior. When a behavior is chosen, it activates the codelets that implement it.

Now turning to Episodic Learrning (EPL) in CTS; the information codelets entered in the WM make it so that an event during consciousness broadcast is considered an episode. These might be learned by CTS's pseudo-hippocampus (PH) in each cognitive cycle. As mentioned above, CTS's pseudo-hippocampus learns the traces. This learning happens through the creation of new sequences of events. Each sequence may contain one or more nodes that have links with other nodes situated in the sequence. Learning happens through the strengthening/weakening of the energy of the nodes and links between them. As regards the information entered in WM, if the PH does not have a response set for the information broadcast by the consciousness broadcast mechanism, it creates a new sequence with a unique ID. It then creates an empty node with a context which explains ongoing situations (current event). Observing all broadcast information by the consciousness mechanism, PH gives a unique ID to each coalition broadcast in the system and saves them instantaneously. To fill out each node, the EM waits for the reasoning phase, the consciously-selected behavior and the ensuing broadcasting of the externally-confirmed event. At this point, each node in the sequence is assigned the time of broadcasted coalition, its total emotional valence, and a key-information-codelet (trigger-codelet) associated to the broadcast coalition that fires the stream of BN (if it has passed its threshold value). The PH then associates the context of the new node with the ID of the broadcast coalition consciously-selected by the Attention mechanism and executed by the BN. The emotional valences corresponding to this broadcast coalition are also saved. At this point the information is ready to be integrated in the different memories of the system. The sequence(s) related to this episode are saved in a database which is considered as CTS's Episodic Memory. This as well as the information learned by CTS's Learning Mechanisms (i.e. learning of regularities, procedural learning and emotional learning during arm manipulation) [6] is distributed and then integrated in the same database separately from Episodic Memory. With this method, CTS clearly relates an episode to its corresponding procedures in the BN. In the next two sections we explain in detail how the episodic memory consolidation and episodic learning processes are implemented in CTS's architecture.

### 4.2 The Episodic Memory Consolidation Model

The memory consolidation process in our cognitive agent takes place after each of CTS's cognitive cycles. Like the human hippocampus, CTS's EPL extracts frequently occurring event sequences during arm manipulation by the astronauts in the virtual world. In our agent, an episodic trace or sequence of events is recorded during *consciousness broadcast* as mentioned in section 3. To mine frequent events sequences, we chose the sequential pattern mining algorithm of [15] which provides several more features than the original GSP sequential pattern mining algorithm [8], such as accepting symbols with numeric values, eliminating redundancy and handling time constraints.

The algorithm takes the database D of all saved sequences of events as input. Here, a sequence of events is recorded for each execution of CTS. An event $X=(i_1, i_2, \ldots i_n)$ contains a set of items $i_1, i_2, \ldots i_n$, and represents one cognitive cycle. For each event, (1) an item represents the coalition of information-codelets that was broadcasted during the cognitive cycle, (2) an optional four items having numeric values indicates the four emotional valences (high threat, medium fear, low threat, compassion) that are associated with the broadcasted coalition, as explained in section 3, and (3) an optional item represents the executed behavior, if one was executed during that cycle. Formally, an events sequence is denoted $s = <(t_1,X_1), (t_2,X_2),\ldots, (t_n,X_n)>$, where each event $X_k$ is annotated with a timestamp $t_k$ indicating the cognitive cycle number.

An events sequence $s_a = <(ta_1,A_1), (ta_2,A_2),\ldots, (ta_n,A_n)>$ is said to be contained in another events sequence $s_b = <(tb_1,B_1), (tb_2,B_2),\ldots, (tb_n,B_m)>$, if there exists integers $1 \leq k1 < k2 < \ldots < kn \leq m$ such that $A_1 \subseteq B_{k1}$, $A_2 \subseteq B_{k2}$, . . . , $A_n \subseteq B_{kn}$, and that $tb_{kj} - tb_{k1}$ is equal to $ta_j - ta_1$ for each $j \in \{1\ldots m\}$. The relative support of a sequence $s_a$ is defined as the percentage of sequences $s \subseteq D$ that contains $s_a$, and is denoted by $supD(s_a)$. The problem of mining frequent events sequences is to find all the sequences $s_a$ such that $supD(s_a) \geq minsup$ for a sequence database D, given a support threshold minsup, and optional time constraints. The optional time constraints are the minimum and maximum time intervals required between the head and tail of a sequence and the minimum and maximum time intervals required between two adjacent events of a sequence. In the experiment described in section 5, we only mined sequences not shorter than 2 time units and not longer than 9 time units, with a maximum number of 2 time unit between any two adjacent events.

Table 1 shows an example of a database containing 6 short sequences. The first event of sequence S1 shows that during cognitive cycle 0, coalition c1 was broadcasted and that an emotional valence of 0.8 for emotion e1 (high threat) was associated with the broadcast. The second event of S1 indicates that at cognitive cycle 1, coalition c2 was broadcasted with emotional valence 0.3 for emotion e2 (medium fear) and that behavior b1 was executed. Table 2 shows some sequences obtained from the application of the algorithm on the database of Table 1 with a minsup of 32 % (2 sequences) and no time constraints. The first frequent pattern is <(0, c1 e1 {0.7})>, which was found in sequences S1, S2, S4 and S6. Because the events containing e1 in these sequences have numeric values 0.8, 0.8, 0.6 and 0.6, the algorithm calculated the average when extracting that pattern, which resulted in the first event having e1 with value {0.7}. Because this pattern has a support of 66 % (4 out of 6 sequences), which is higher than minsup, it is deemed frequent.

**Table 1. A Data Set of 6 Sequences**

| ID | Events sequences |
|---|---|
| S1 | <(0, c1 e1{0.8}), (1, c2 e2{0.3}) b1)> |
| S2 | <(0, c1 e1{0.8}), (1, c3), (2, c4 b4), (3, c5 b3)> |
| S3 | <(0, c2 e2{0.3}), (1, c3), (2, c4), (3, c5 b3)> |
| S4 | <(0, c3), (1, c1 e1{0.6} b4),(2, c3)> |
| S5 | <(0, c4 b4), (1, c5), (2, c6)> |
| S6 | <(1, c1 e1{0.6} b4), (2, c4 b4), (3, c5)> |

Finally, among mined sequences, there can be many redundant sequences. To facilitate consolidation of memory and decrease redundancy, we choose to mine only frequent closed sequences. "Closed sequences" are the sequences not contained in any other sequence with the same support. A closed pattern induces an equivalence class of patterns sharing the same closure, i.e. all the patterns belonging to the equivalence class are verified by exactly the same set of sequences (they have the same support). The set of closed frequent sequences is a compact representation of the set of frequent sequences, because it allows the reconstitution of the set of all frequent sequences and their support [14]. In this work, we choose to mine only closed sequences as it results in fewer sequences, and prevents information loss.

### 4.3 The Episodic Learning Process

The episodic learning mechanism constantly adapts to the agent behavior by intervening in the coalitions selection phase of CTS, as follows. At each cognitive cycle, before choosing the candidate coalition to be broadcasted, all the frequent events sequences are scanned for finding those that matches with the coalitions broadcasted at previous cognitive cycles. The last n broadcasted coalitions are represented as a sequence Sc=<(0,c$_1$), (1,c$_2$)... (n,c$_n$)>. The "*episodic learning*" mechanism checks all frequent sequences to find each sequence Sa=<(t$_1$,a$_1$), (t$_2$, a$_2$), ... (t$_n$,a$_n$)> such that for a number k > 1, the sequence defined by the last k broadcasts of Sc <(t$_{n-k+1}$, c$_{n-k+1}$), (t$_{n-k+2}$,c$_{n-k+2}$) ... (t$_n$,c$_n$)> is included in Sa and there is an event following that sequence occurrence that contains a coalition that is candidate for being broadcast during the current cognitive cycle. The "episodic learning mechanism" then computes, for each sequence, the sequence strength which is defined as the support of the sequence multiplied by the sum of emotional valences associated to each broadcast occurrence in the sequence. The episodic learning mechanism then adds, for each coalition that is candidate to be broadcast, the strength of the strongest sequence containing it and the strength of the weakest sequence containing it. The "episodic learning mechanism" selects the coalition with the highest sum. This coalition will be broadcast in favor of other coalitions waiting to be broadcasted.

**Table 2. Example of Events Sequences Extracted**

| Mined sequences | Support |
|---|---|
| <(0, c1 e1{0.7})> | 66 % |
| <(0, c3), (2, c5 b3)> | 33 % |
| <(0, c4 b4), (1, c5)> | 50 % |
| <(1, c3), (2, c4), (3, c5 b3)> | 33 % |
| … | … |

## 5. EXPERIMENTATION

We performed various preliminary experiments to validate CTS's new episodic learning. One experiment conducted with user 3 is here detailed.

User 3 tended to make frequent mistakes when he was asked to guess the arm distance from a specific part of the ISS and its name (Figure 2.A). Obviously, this situation caused collision risks between the arm and ISS and was thus a very dangerous situation. This situation was implemented in the CTS's Behavior Network [14] (Figure3). In this situation, the CTS Episodic Learning mechanism (Figure 2.B) had to make a decision between giving a direct solution "*You should move joint SP*" (Figure 3. Scenario 1) or a brief hint such as "*This movement is dangerous. Do you know why?*" (Figure 3. Scenario 2). Interacting with different users, CTS learned about their profiles. For user 3, the experiments showed that it is better to first give a hint and then ask the user to think about the situation and lastly giving him the answer. At the end of each scenario CTS asks an evaluation question to verify the effectiveness of its interventions (Figure 3. Evaluation). As we mentioned previously, CTS Behavior Network (Figure 3) possesses two paths to solve this problem. In the case of User 3, CTS chose the second option and presented a hint followed by an answer (Figure2.C). User 3 did not remember the answer, so a help (a hint and not a solution) was proposed to him to push him to think more about the situation. He still did not remember, so CTS finally signaled to User 3 that this situation would end with a collision with the ISS.

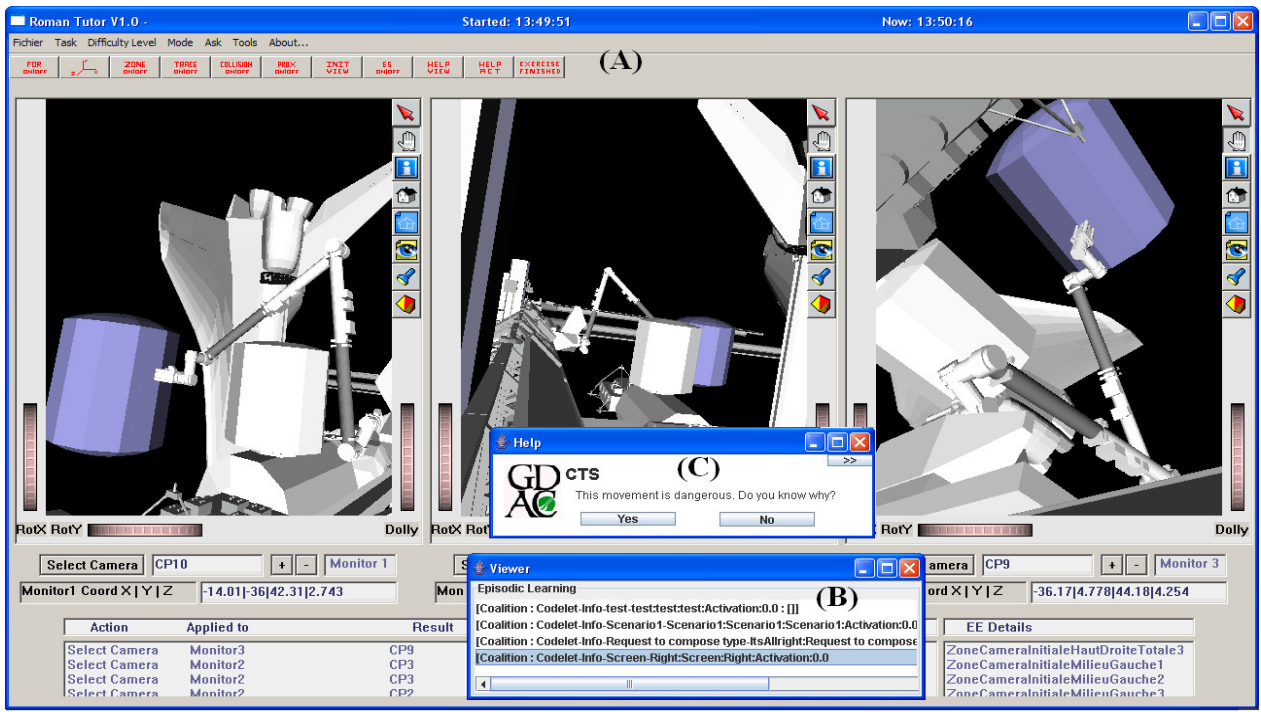

**Figure 2. (A) Simulator interface (B) Episodic Learning Viewer (C) CTS Intervention**

During the interaction with this user, CTS learned the valences of each trace (Figure 3. Scenario1 and 2) with its nodes and all emotional valences being saved at the end of each execution. The average length of the stored sequences was of 26 events. One partial trace saved when CTS gave a hint (scenario 2) to User 3 was <(13, c11), (14, c14), (15, c15), (16, c18), (17, c19 e4 {0.8})>. In this trace, the positive valence 0.8 for emotion e4 (compassion) was recorded because the learner answered to the Evaluation question correctly after receiving the hint. Thus during CTS's interaction with User 3, when User 3 gave the correct answer to CTS, the EM gave more positive valences to the scenario which might have been chosen in the future by EPL mechanism. In another partial trace saved by CTS, when User 3, having received some help, answered incorrectly to a question (scenario 1) <(16, c11), (17, c14), (18, c16), (19, c17), (20, c20 e2 {-0.4})>, CTS associated the negative valence -0.4 to emotion e2 (medium fear). After five executions, the memory consolidation phase extracted ten frequent event sequences, with a minimum support (minsup) higher than 0.25. This way, had the CTS have had to face the same problem in the future, it might have chosen between scenario 1 and 2.

The episodic learning process evaluates all consolidated patterns and tries to detect the best pattern for each situation having ended by self-satisfaction (OCC model) in CTS. CTS chooses the pattern < (0, c11), (1, c18), (3, c18), (4, c19 e4 {0.8})>, because it contains the most positive emotional valence, has the highest frequency, and many of its events match with the current events executed in CTS. In our model, CTS will thus choose the path which gives it greatest self-satisfaction (scenario 2). When the emotional valence is not as positive as was the case in User 3, CTS might choose scenario 1 rather than scenario 2. Moreover, because the set of patterns is regenerated after each CTS cognitive cycle, some new patterns are created and others disappear. This may cause some changes in CTS's behavior.

This suggests that CTS's architecture now adapts better to any situation and dynamically learn how to associate an episode to one or more scenarios in the BN. It also suggests that it now learns how to choose the path that is the most probable of bringing self-satisfaction.

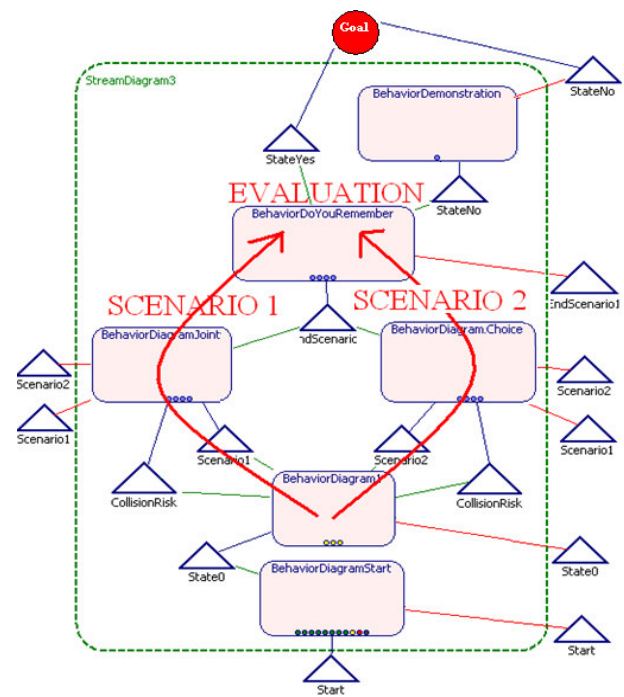

**Figure 3. Part of the CTS Behavior Network**

## 6. Behavior of the Data mining Algorithm in Mining Patterns from Recorded Sequences

Here, we sought to observe how the data mining algorithm behaved when the number of recorded sequences increased. The experiment was done on a 3.6 GHz Pentium 4 desktop computer running Windows XP, and consisted of performing 160 CTS executions for a situation similar to the one described above. In this type of situation, CTS conducts a dialogue with the student that includes from two to nine messages or questions (an average of six) depending on what the learner answers and the choices CTS makes (situation similar to that of choosing between scenarios 1 and 2).

During each trial, we randomly answered the questions CTS asked. Each recorded sequence contained approximately 26 broadcasts. Figure 4 presents the results of the experiment. The first graph shows the time required for mining frequent patterns after each CTS execution. From this graph, we see that the time for mining frequent patterns was generally short (not more than 6 seconds) and that the algorithm's behavior tended to increase linearly with the number of recorded sequences. For our purposes, this performance is promising. However, it could still be improved for we have not yet fully optimized the algorithm to perform incremental mining of sequential patterns as some other sequential patterns mining algorithms do [20]. We are currently working on making it possible, for it would improve performance, as it would no longer be necessary to recompute the set of patterns for each new added sequence from scratch. The second graph shows the average size of patterns found during each execution. It ranges from 9 to 16 broadcasts. The third graph depicts the number of patterns found. It remained low and stabilized at around 8.5 patterns during the last executions. The reason why the number of patterns is small is that we mined only closed patterns (c.f. section 4.2). If we had not mined only closed patterns, the number of patterns would have been much higher as all the subsequences of all patterns would have been included in the results. Mining closed patterns is also much faster as during the search for patterns large parts of the search space that are guaranteed not to lead to close patterns are pruned. For example, for mining non closed patterns from the first four sequences only, it took more than 1 hour (we stopped the algorithm after 1 hour), while mining closed patterns took only 0.558 second. The reason for this is that the four sequences share more than fifteen common broadcasts. Therefore, if the pruning of the search space is not done, the algorithm has to consider all combinations of these broadcasts, which is computationally very expensive. This demonstrates that it is beneficial to mine closed patterns. Finally, the fourth graph presents the average time for executing the Episodic learning algorithm at each execution. This time was always less than 5 milliseconds. Thus, the costliest operation of the learning mechanism is definitely the extraction of patterns.

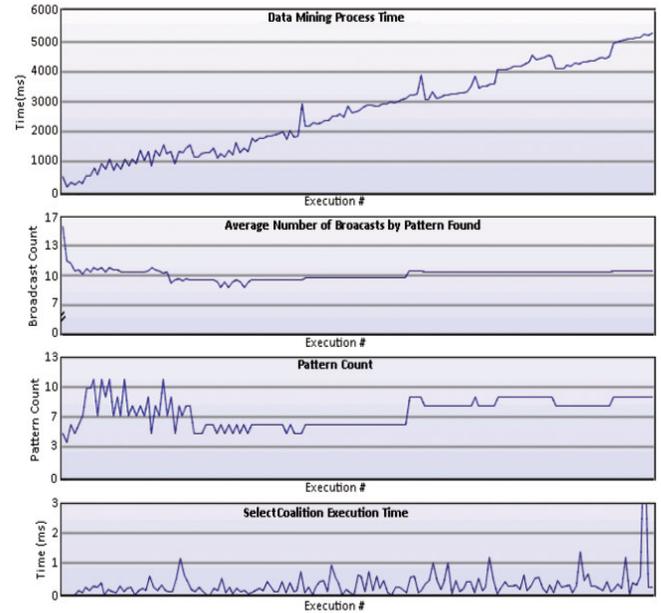

Figure 4: Results from second experiment

## 7. CONCLUSION

As far as we know, no cognitive agent presently uses data mining algorithms. Nonetheless, they have proven very useful for extracting significant information from the huge amount data that they have to handle. The interaction between an agent and its dynamic environment also generates large amounts of data. The episodic learning algorithm used in this work is inspired from a memory consolidation theory which is biologically plausible. The collaboration between the Emotional mechanism and this Episodic Learning helps to choose the corresponding behaviors that are most likely to bring the agent to a self-satisfactory emotional state. CTS learns, during real time interactions with astronauts, how to associate an event and its corresponding emotional valences witha partial/complete sequences of behaviors chosen by their Behavior Network for execution.

In the future, we will perform further experiments to measure empirically how CTS influence the learning of students. We will investigate different ways of improving the performance of our sequential pattern mining algorithm, including the possibility of modifying it to perform an incremental mining of sequential patterns. We also plan to compare the Episodc Learning mechanism with others agent learning mechanisms. Lastly, we are interested in exploring different forms of learning in multi-agent systems by applying data mining techniques, mining other types of temporal patterns such as trends for improving agents' behavior, and mining patterns from group behaviors.

## 8. ACKNOWLEDGMENTS

Our special thanks to our colleague Sioui Maldonado Bouchard for her collaboration in this paper, and Daniel Dubois, Sebastien Dubourdieu and Mohamed Gaha for their contribution to the development of CTS. The authors also thank the Fonds Québécois de la Recherche sur la Nature et les Technologies and the Natural Sciences and Engineering Research Council for their financial support and DR.jean-Yves Housset for our discussions and pertinent suggestions.